\documentclass[runningheads]{llncs}
\pdfoutput=1

\PassOptionsToPackage{hidelinks}{hyperref}
\usepackage[T1]{fontenc}
\usepackage{graphicx}
\usepackage{amsmath}
\usepackage{booktabs}
\usepackage{array}
\usepackage{xcolor}
\usepackage{orcidlink}   

\begin{document}

\title{HERMES: A Hybrid Ensemble for Head-and-Neck Tumor Segmentation, TN Staging, and Recurrence-Free Survival on PET/CT}

\titlerunning{HERMES: Segmentation, Staging, and Survival on H\&N PET/CT}

\author{Kai Wang\inst{1}\orcidlink{0000-0003-2155-1445}\thanks{Corresponding author: \email{kai.2.wang@cuanschutz.edu}} \and
Meixu Chen\inst{1}\orcidlink{0000-0003-3095-1256} \and
Elie Nasr\inst{1}\orcidlink{0009-0003-7881-870X} \and
Ryan Lanning\inst{1}\orcidlink{0000-0002-9267-519X} \and
Moyed Miften\inst{1}\orcidlink{0009-0008-3619-2768}}

\authorrunning{K. Wang et al.}

\institute{Department of Radiation Oncology, University of Colorado School of Medicine, Aurora, CO, USA \\
\email{\{kai.2.wang,meixu.chen,elie.nasr,ryan.lanning,moyed.miften\}@cuanschutz.edu}}

\maketitle

\begin{abstract}
We present \textbf{HERMES} (Hybrid Ensemble for Radiotherapy-target segmentation, Malignancy staging, and Event-free Survival), a single containerized algorithm for the three HECKTOR 2026 subtasks: segmentation of the primary tumor (GTVp) and lymph nodes (GTVn), radiological T/N staging, and recurrence-free survival (RFS) prediction, computed from a paired FDG-PET/CT scan and an electronic health record. A 10-fold ensemble of STU-Net Small networks produces the segmentation, and the predicted mask then drives two downstream tasks. Rather than pass a generic radiomics vector to the staging models, we derive from the predicted masks a compact set of geometry features (nodal count, size, and burden) aligned with the size and number axes of AJCC/UICC 7th-edition radiological T/N staging. On internal cross-validation these features raise N-stage balanced accuracy from 0.691 to 0.720 at lower feature dimensionality (a favorable trend whose paired interval includes zero); read from ground-truth masks they reach 0.897 and beat radiomics by a statistically significant margin, indicating the gain reflects mask-derived geometry rather than a mask artifact and is diluted at deployment by imperfect segmentation. For prognosis we combine complementary deep and clinical risk experts in an equal-weight ensemble, one trained with a concordance-tracking survival loss whose value tracks the concordance index during training. Every component is selected on out-of-fold cross-validation under a regularization-oriented protocol. On the HECKTOR 2026 validation set, HERMES attained a Mean Dice of 0.641, T and N balanced accuracies of 0.580 and 0.642, and an RFS C-index of 0.679, and qualified for the testing phase. \textbf{Team: AMC\_HNC.}

\end{abstract}

\section{Introduction}

The HECKTOR 2026 challenge targets the clinical decision chain for head-and-neck (H\&N) oropharyngeal cancer from FDG-PET/CT: delineation of the primary tumor (GTVp) and involved lymph nodes (GTVn), radiological T and N staging, and recurrence-free survival (RFS) prediction~\cite{ref1,ref2}. Unlike previous editions that scored separate algorithms, HECKTOR 2026 requires a single containerized algorithm emitting all three outputs, reflecting a clinical workflow in which a delineation informs staging and outcome estimation. Prior HECKTOR editions established PET/CT GTVp/GTVn segmentation and outcome prediction as a public benchmark~\cite{ref1,ref2,ref3}; segmentation entries have converged on U-Net-family encoders~\cite{ref4} with heavy augmentation, with nnU-Net~\cite{ref5}, the scalable STU-Net~\cite{ref6}, and transformer variants~\cite{ref7} anchoring recent results, usually combined into multi-fold, multi-augmentation ensembles with test-time augmentation, and increasingly with an FDG-uptake-guided region-of-interest localization step that the strongest recent entries share~\cite{ref2}. For staging and survival, radiomics with clinical variables remains a strong, data-efficient baseline~\cite{ref8,ref9}, mask-guided survival prediction has been explored in prior HECKTOR editions~\cite{ref10}, and deep survival networks commonly use the Cox partial likelihood (DeepSurv-style)~\cite{ref11,ref12} or discrete-time and ranking objectives~\cite{ref13}.

Two observations shaped our approach. First, staging and prognosis are computed downstream of segmentation, so the pipeline depends above all on how well the predicted mask is converted into task-relevant signal. This motivates our central design choice: to summarize the mask with features that mirror the definitions of the downstream tasks rather than with a generic descriptor. Generic radiomics captures intensity and texture, but for nodal staging, whose categories are defined by node number, size, and laterality, an intensity/texture vector captures none of these axes directly. Second, on a small public validation set (roughly 50 patients), honest out-of-fold (OOF) cross-validation is a more reliable guide than the leaderboard, because the validation-to-test gap is where over-fitted models regress. We therefore select every component on OOF cross-validation, prefer strongly-regularized and equal-weight-ensembled models, and do not tune on the public validation set.

\begin{figure}[t]
\centering
\includegraphics[width=\textwidth]{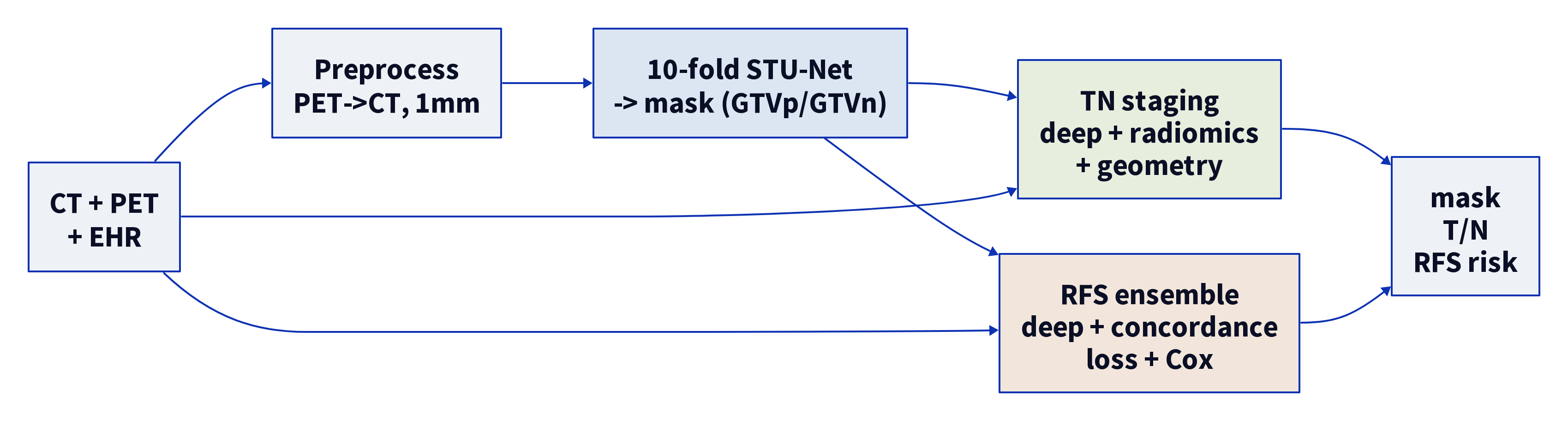}
\caption{HERMES pipeline. FDG-PET/CT and EHR feed a 10-fold STU-Net segmentation; the predicted mask then drives TN staging (deep, radiomics, and geometry experts) and RFS (a deep survival ensemble with a concordance-loss expert plus a clinical Cox model).}
\label{fig:pipeline}
\end{figure}

We realize these choices in \textbf{HERMES} (Hybrid Ensemble for Radiotherapy-target segmentation, Malignancy staging, and Event-free Survival), our single containerized entry. Its central scientific contribution is (1) \textbf{clinically meaningful geometry features for staging}: compact descriptors read from the predicted masks and aligned with the size and number axes of 7th-edition radiological T/N staging, which outperform a 30-dimensional radiomics block for N-staging at lower dimensionality, an effect we further probe against ground-truth masks to show it reflects the geometry signal rather than a mask-quality artifact. The remaining three are engineering and protocol choices that make this finding deployable and robust rather than novel in themselves: (2) a unified, single-container multi-task pipeline (Fig.~\ref{fig:pipeline}) in which a 10-fold STU-Net ensemble drives predicted-mask-based staging and survival; (3) a survival expert trained with a \textbf{concordance-tracking loss}~\cite{ref14}, adopted for its interpretable, metric-aligned training signal rather than for a decisive accuracy gain; and (4) a regularization-oriented selection protocol appropriate to a small, high-variance validation set. Throughout, we report where design gains are within cross-validation noise.

\section{Materials and Methods}

\textbf{Data.} The HECKTOR 2026 training set comprises 782 patients from 8 centers (CHUM 56, CHUP 75, CHUS 72, CHUV 51, HGJ 55, HMR 18, MDA 444, USZ 11); the full challenge cohort, with hidden multi-center test data including previously unseen centers, is larger. T/N labels are the challenge-provided radiological stages under AJCC/UICC 7th edition~\cite{ref15}; we do not define or annotate staging ourselves. For RFS, 727 patients have valid survival labels with 148 events (an event rate near 20\%). Each case is provided as a native-grid CT, a native-grid PET in SUV units, and an \texttt{ehr.json}. Segmentation is scored by per-patient Mean Dice over GTVp and GTVn, staging by balanced accuracy (mean per-class recall), and survival by Harrell's concordance index~\cite{ref16,ref17}.

\textbf{Preprocessing.} PET is resampled onto the CT voxel grid (SimpleITK linear), applied once so that segmentation, radiomics, and the deep patch model operate on aligned channels; volumes are reoriented to RAS and resampled to 1\,mm isotropic. CT is windowed to $[-200,200]$\,HU and rescaled to $[0,1]$, and PET is percentile-clipped to $[0.5,99.5]$ and rescaled to $[0,1]$ for the network input, a scale-invariant normalization robust to whether the PET intensity scale is SUV or Bq/ml. A PET-intensity region-of-interest crop anchors the field of view on the high-uptake tumor region, following the FDG-guided localization strategy that strong entries in prior HECKTOR editions have converged on~\cite{ref2,ref3}. Clinical variables form an 18-dimensional encoding (age, gender, HPV status, tobacco, alcohol, performance status, treatment); the linear models use a 13-dimensional subset that drops one reference level per categorical group.

\textbf{Training and validation setup.} All components are developed and selected by cross-validation on the 782-patient training cohort: 10-fold for segmentation and 5-fold for the staging and survival models, where the 10-fold split refines the same patient partition, reporting pooled OOF metrics. Selected components are re-fit on the full training set for the deployed container. Model selection uses only OOF cross-validation; no component is tuned on the organizer's held-out public validation set (about 50 patients), which serves solely as an independent final assessment (Sec.~\ref{sec:results}). Networks were built with MONAI~\cite{ref18} and trained on RTX-class GPUs; the deployed pipeline runs within the challenge's NVIDIA T4 GPU (16\,GB) and 25-minute-per-case budget. The segmentation networks are trained from random initialization (no pretrained warm-start) at 1\,mm isotropic resolution on $160\times160\times192$ patches with deep supervision, a plateau-decayed learning rate, and early stopping, and the three highest-validation-Dice checkpoints per fold enter the ensemble. The deep staging and survival models use AdamW (weight decay $10^{-4}$), a cosine-annealed learning rate of $3\times10^{-4}$, batch size 8, and up to 80 epochs with early stopping on the validation task metric (C-index for survival, balanced accuracy for staging). All settings are collected in Table~\ref{tab:hyper}.

\subsection{Segmentation}
\label{sec:seg}

We use STU-Net Small (about 14\,M parameters)~\cite{ref6}, a scalable residual U-Net reported to match or exceed nnU-Net on large-scale benchmarks, which we adopt as our single segmentation architecture; it maps the 2-channel CT+PET volume to a 3-class (background/GTVp/GTVn) logit map and is trained with a Dice plus cross-entropy loss and deep supervision. Training uses 10-fold cross-validation with an nnU-Net-style augmentation profile~\cite{ref5}: per-modality low-resolution simulation (probability 0.25, for unseen-center resolution variability), Gaussian noise and smoothing, brightness/contrast/gamma, three-axis flips, affine and elastic transforms, and random PET-channel dropout (probability 0.1). The deployed ensemble comprises 30 checkpoints, the three best per fold across the 10 folds, each evaluated with test-time augmentation (identity, three axis flips, mild gamma); softmax probabilities are averaged before a single argmax, with checkpoints streamed one at a time to bound memory. A per-class connected-component (CC) filter removes GTVp components below 1000\,mm$^3$ and GTVn components below 500\,mm$^3$ and keeps the two largest GTVp and eight largest GTVn components, and a PET SUV gate removes GTVn components whose peak value in the original SUV-unit PET (as supplied by the challenge) falls below 2.5, a threshold defined in absolute SUV rather than on the normalized network input; these post-processing thresholds were fixed on OOF. The predicted mask produced here is the geometric input shared by the staging and survival modules that follow, so segmentation quality is the common upstream dependency of both downstream tasks.
\subsection{TN Staging with Geometry Features}
\label{sec:staging}

Staging reuses the predicted mask through two complementary experts and, for the enhanced configuration, a set of geometry features. The deep expert is a 3-D ResNet-18~\cite{ref19} (\texttt{DualHeadFusionResNet}, base width 32) over a $96^3$ CT+PET patch centered on the predicted-mask centroid; the predicted mask is one-hot encoded and fused early through a small convolutional branch, and the pooled feature is concatenated with an MLP embedding of the clinical vector before T and N heads. The classical expert is a 30-dimensional first-order/shape radiomics descriptor~\cite{ref8,ref20} computed on the predicted mask (per-ROI intensity statistics for CT and PET, volume, and total-lesion-glycolysis) concatenated with the clinical block, implemented self-contained in NumPy/SciPy/SimpleITK.

Our staging contribution replaces the radiomics block with features that instantiate the staging criteria. In AJCC/UICC 7th-edition radiological staging, nodal category depends on the number, size, and laterality of involved nodes (a single ipsilateral node no larger than 3\,cm is N1, multiple ipsilateral nodes are N2b, and any node larger than 6\,cm is N3), and primary category on tumor size for T1--T3. From the CC-filtered predicted GTVn (26-connectivity, components no smaller than 500\,mm$^3$) we compute the nodal component count, the largest-component volume and axis-aligned extent, the dominance fraction (largest component over total nodal volume), and the total nodal burden, together with log-transformed volume terms; from the predicted GTVp we compute the primary-tumor extent and volume.

Two caveats accompany the features. First, they capture number and size but not laterality, one of the three N-staging axes, and they use an axis-aligned bounding-box extent, which under-estimates the true greatest dimension for obliquely-oriented nodes. Second, they are computed on imperfect predicted masks (GTVn Dice near 0.69) and from connected components, which merge conglomerate nodal masses; the monotone stage trend is empirical rather than a claim of exact agreement with reference measurements.

The decision rules are fixed on OOF. The deep model is trained with both T and N heads, but on OOF the N-head does not improve over the geometry model and is therefore not used in the deployed N decision, whereas the T-head contributes to T staging through fusion; this asymmetry motivates the mask-derived geometry route for N. N-stage is an L1-regularized logistic regression on clinical and geometry features (regularization $C=0.03$, chosen for stability), replacing the 30-dimensional radiomics block with the 7-dimensional geometry block. T-stage is an equal-weight fusion of the deep T-head and a logistic regression over clinical, radiomics, and primary-geometry features, with the argmax restricted to T1--T4. The resulting comparisons are reported in Sec.~\ref{sec:results}.
\subsection{Recurrence-Free Survival}
\label{sec:rfs}

Like staging, RFS is computed downstream of the predicted mask: its deep experts read CT+PET patches centered on the mask centroid. RFS is an equal-weight, z-averaged ensemble of complementary risk experts. Each expert produces a scalar risk (higher meaning higher hazard) that is z-normalized using training-fold statistics only, so that no test information leaks through normalization, and the normalized risks are averaged. Equal weighting is a deliberate anti-overfitting choice: with about 700 survival-labeled patients, tuning fusion weights over-fits, whereas equal weighting is a stable, parameter-free combination. The members are a multi-scale deep survival model (\texttt{triplehead}, a single member evaluated at $96^3$ and $112^3$ patches), two single-task deep models (\texttt{rfs\_only}, one with additional affine augmentation) for diversity, the concordance-loss expert below, and a classical risk model, giving four members in HERMES-base and five in HERMES+.

The Cox partial-likelihood value does not track ranking quality, so it is an unreliable signal for early stopping and checkpoint selection. We instead train one deep expert with a concordance-tracking loss~\cite{ref14}, a smooth pairwise surrogate over comparable pairs $\mathcal{P}=\{(i,j):\delta_i=1,\ t_j>t_i\}$ in which subject $i$ experiences the event and subject $j$ outlives it:
\begin{equation}
\mathcal{L} \;=\; \frac{1}{|\mathcal{P}|}\sum_{(i,j)\in\mathcal{P}} \sigma\!\Big(-\tfrac{r_i - r_j}{\tau}\Big), \qquad \tau = 0.1,
\end{equation}
where $r$ are the predicted risks and $\sigma$ the logistic function. The summand is near 1 when a pair is mis-ranked and near 0 when correctly ranked, so $\mathcal{L}$ approximates $1-\text{C-index}$ and falls with concordance throughout training. We train one deep expert by substituting this loss for the Cox likelihood, keeping architecture, data, batch size (8), and schedule identical; Sec.~\ref{sec:results} reports the resulting comparison. For the classical expert we prefer a pure-clinical ridge Cox over a radiomics LASSO-Cox (tied in OOF) for the enhanced configuration, because radiomics is sensitive to scanner and protocol~\cite{ref20,ref21} and transfers less reliably to unseen centers, whereas clinical variables transfer directly.
\subsection{Model Selection}
\label{sec:selection}

Every component is selected on leakage-free OOF predictions (features and risks from a model that never saw the patient, OOF-predicted masks, training-fold normalization), ranked by the pooled OOF metric rather than the fold-lottery-inflated per-fold mean, and re-fit on all data for deployment. We prefer the more regularized, more stable option even at a marginal cost in point estimate (stronger L1 for N-staging, equal-weight rather than tuned fusion, clinical rather than radiomics-heavy risk for cross-center transport). We report a baseline (\textbf{HERMES-base}) and an enhanced variant (\textbf{HERMES+}) that adds the geometry staging features, the concordance-loss survival expert, and the clinical-transport risk model; their comparison isolates these components.

\begin{table}[t]
\centering
\caption{Key hyperparameters.}
\label{tab:hyper}
\small
\renewcommand{\arraystretch}{1.1}
\begin{tabular}{p{3.1cm}p{8.3cm}}
\toprule
Component & Setting \\
\midrule
Segmentation & STU-Net Small (about 14\,M), random init; 10-fold; Dice+CE with deep supervision; 1\,mm iso, $160{\times}160{\times}192$ patch; nnU-Net augmentation; plateau LR, early stop; deploy: 30 ckpts (top-3/fold by val Dice) with flip/gamma TTA, streamed \\
Post-processing & CC filter GTVp $\ge$1000\,mm$^3$ (top-2), GTVn $\ge$500\,mm$^3$ (top-8); nodal SUV gate 2.5 \\
Deep staging/survival & DualHeadFusionResNet (3-D ResNet-18, width 32; $96^3/112^3$ patch, mask branch, 18-d clinical MLP); AdamW (wd $10^{-4}$), cosine LR $3\times10^{-4}$, batch 8, $\le$80 epochs, early stop \\
Concordance loss & pairwise sigmoid over comparable pairs, temperature $\tau=0.1$ \\
N-stage & L1 logistic regression, $C=0.03$; features: 13-d clinical, 7-d nodal geometry \\
T-stage & equal-weight fusion of deep T-head and logistic regression (clinical, radiomics, primary geometry), argmax over T1--T4 \\
RFS ensemble & equal-weight z-average of triplehead ($96^3/112^3$), rfs\_only, affine-aug rfs\_only, concordance-loss expert, and a classical Cox (LASSO base, clinical ridge for HERMES+) \\
Clinical encoding & 18-d (13-d for linear models) \\
\bottomrule
\end{tabular}
\end{table}

\section{Results}
\label{sec:results}

\textbf{Validation performance.} On the HECKTOR 2026 validation set, the deployed HERMES+ configuration attained the per-task scores in Table~\ref{tab:val} and qualified for the testing phase. Two evaluation settings appear below and should not be conflated: Table~\ref{tab:val} is the organizer's held-out validation set (about 50 patients, hidden labels), whereas all ablations (Table~\ref{tab:abl}) are internal pooled out-of-fold (OOF) results on the 782-patient training cohort, with patient-level bootstrap 95\% confidence intervals (2000 resamples). We treat OOF as primary for model selection because the roughly 50-patient validation set is high-variance: three tasks sit below their OOF values (validation N 0.642, segmentation 0.641, RFS 0.679 versus OOF 0.720, 0.701, 0.715) and T-stage sits above (0.580 versus 0.454), a mix of mild OOF optimism and small-sample variance that we read cautiously. Because the validation labels are hidden, the geometry-versus-radiomics comparison cannot be repeated there and is reported on OOF only; the testing leaderboard is likewise hidden until the event.

\begin{table}[t]
\centering
\caption{HERMES+ (deployed) validation-phase performance on the organizer's held-out set ($\sim$50 patients); internal cross-validation appears in Table~\ref{tab:abl}.}
\label{tab:val}
\begin{tabular}{lc}
\toprule
Metric & Score \\
\midrule
Segmentation, Mean Dice & 0.641 \\
T-stage, balanced accuracy & 0.580 \\
N-stage, balanced accuracy & 0.642 \\
RFS, C-index & 0.679 \\
\bottomrule
\end{tabular}
\end{table}

\begin{figure}[t]
\centering
\includegraphics[width=0.58\textwidth]{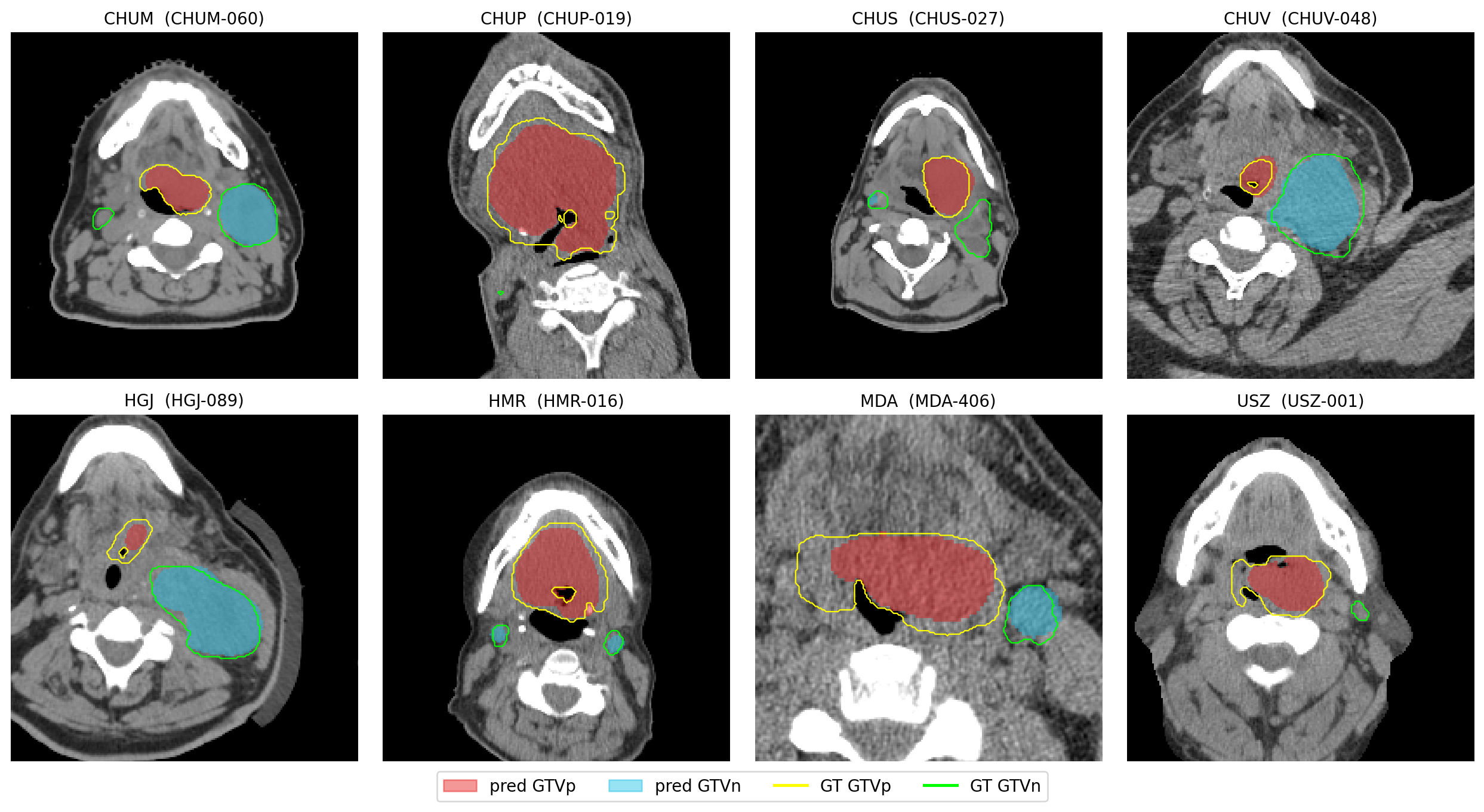}
\caption{Predicted (filled: red GTVp, cyan GTVn) versus ground-truth (contours: yellow GTVp, green GTVn) segmentation on the maximum-tumor axial slice for one representative patient from each of the eight training centers (deployed-model out-of-fold predictions).}
\label{fig:overlays}
\end{figure}

\textbf{Segmentation.} The deployed 30-checkpoint ensemble attains per-patient Mean Dice 0.714 (GTVp), 0.688 (GTVn), and 0.701 (overall) on OOF masks. A 10-fold ensemble marginally exceeds a 5-fold one (0.701 versus 0.697), within cross-validation noise, and we retain it for the greater test-time diversity. The post-processing is fixed on OOF: a grid sweep of per-class connected-component thresholds is optimal at GTVp 1000\,mm$^3$ and GTVn 500\,mm$^3$, and disabling CC filtering entirely lowers the aggregated segmentation score (a composite of per-patient Dice and lesion detection) by about 0.013 relative to this optimum. The nodal SUV gate raises aggregated GTVn lesion-detection F1 from 0.686 to 0.699 (precision 0.70 to 0.73) while per-patient Mean Dice is unchanged, confirming that it removes low-uptake nodal false positives at no overlap cost. Predictions are visually consistent with the reference across all eight centers, including the two smallest cohorts, HMR and USZ (Fig.~\ref{fig:overlays}); no center fails systematically, which is reassuring given that the hidden test set includes previously unseen centers.

\begin{figure}[t]
\centering
\includegraphics[width=0.58\textwidth]{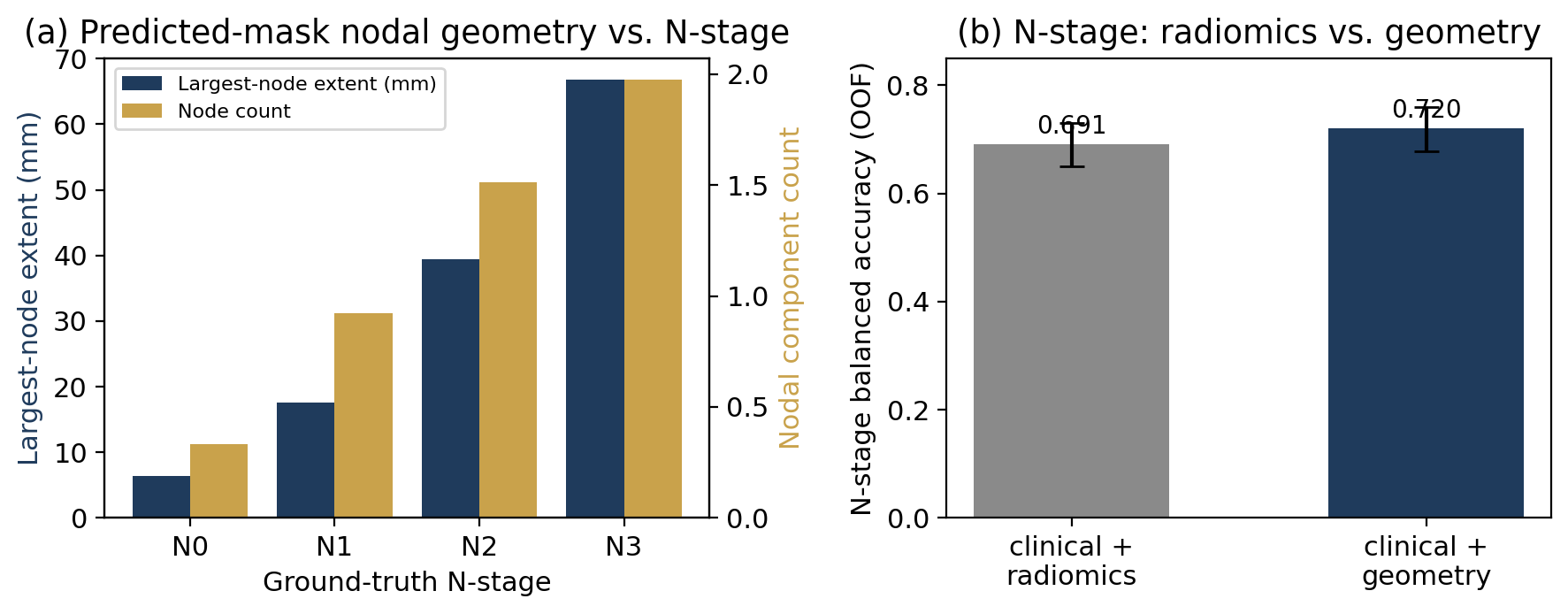}
\caption{(a) Predicted-mask nodal geometry (largest-node extent, node count) is monotone in ground-truth N-stage. (b) N-stage balanced accuracy (out-of-fold, patient-level bootstrap 95\% CIs): clinical plus geometry (0.720) versus clinical plus radiomics (0.691).}
\label{fig:geom}
\end{figure}

\textbf{TN staging.} The nodal geometry features are monotone in the ground-truth stage: mean largest-component extent rises from 6.3\,mm at N0 to 66.7\,mm at N3 and mean nodal count from 0.3 to 2.0 (Fig.~\ref{fig:geom}a), confirming that the predicted mask carries the staging signal. Replacing the 30-dimensional radiomics block with the 7-dimensional geometry block raises N-stage balanced accuracy from 0.691 (95\% CI [0.650, 0.730]) to 0.720 ([0.678, 0.760]), a paired gain of $+0.030$ (95\% CI [$-0.008$, $+0.067$]) at lower dimensionality (Fig.~\ref{fig:geom}b). The interval includes zero, so we do not claim significance, though 0.95 of the paired bootstrap difference lies above zero. Several further checks make it credible. Dropping radiomics helps: only geometry terms (total burden, largest-node extent, dominance) and one clinical variable survive L1 selection, so intensity and texture are largely uninformative for nodal category once size and number are available. The effect is stable across regularization strength (0.712 to 0.721 for $C\in\{0.03,0.05,0.1\}$), whereas a random forest on the same features is markedly worse (0.61), consistent with our preference for the simpler, regularized model. Most tellingly, a two-by-two of feature type against mask source (the ground-truth-mask rows of Table~\ref{tab:abl}) separates feature quality from mask quality. On ground-truth masks geometry reaches 0.897 and radiomics 0.837 (each at its own best regularization), a paired gain of $+0.060$ (95\% CI [$+0.028$, $+0.093$]) that, unlike the predicted-mask gap, excludes zero: with mask noise removed, geometry beats radiomics as features significantly, so the predicted-mask advantage is a real effect diluted by imperfect segmentation, not an artifact of it. Both feature sets also gain sharply from ground-truth masks, so segmentation quality is the shared bottleneck and the deployed number sits below a perfect-mask ceiling. For T-staging, adding primary-tumor geometry to the deep and radiomics fusion moves balanced accuracy from 0.444 to 0.454, within noise; T remains the weakest task for the reason discussed in Sec.~\ref{sec:discussion}.

\begin{figure}[t]
\centering
\includegraphics[width=0.92\textwidth]{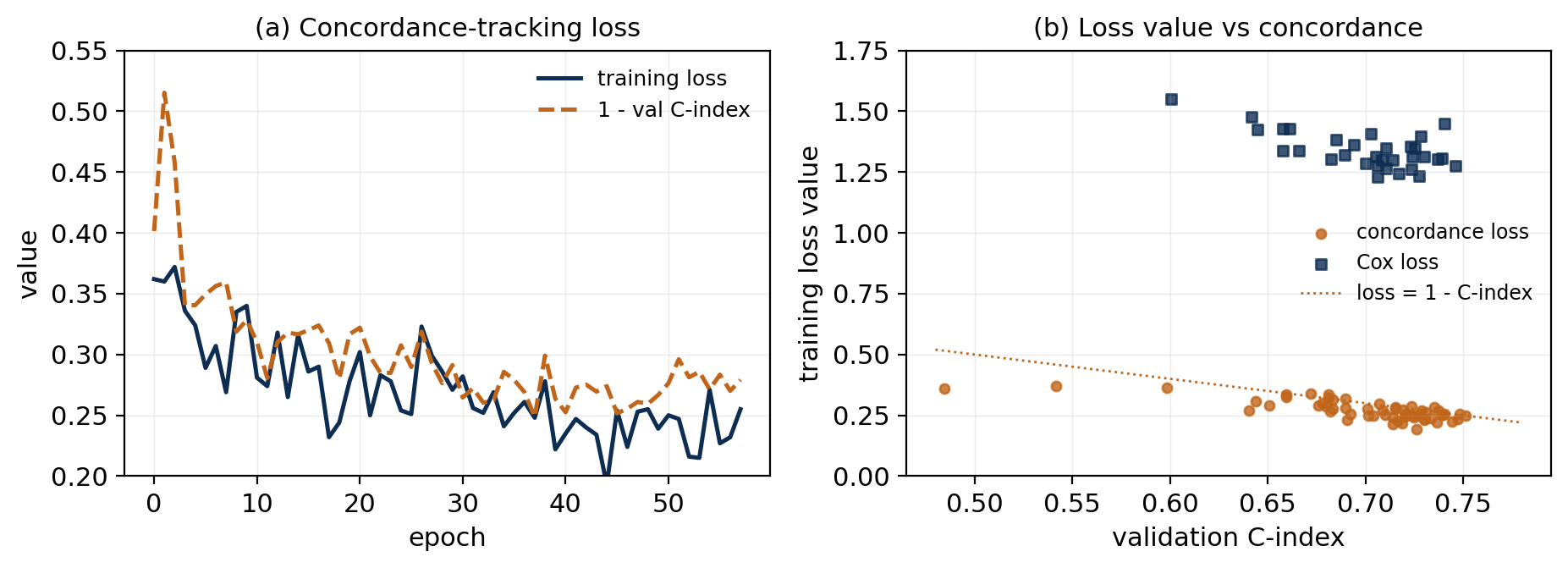}
\caption{The concordance loss value is an interpretable, metric-aligned training signal (representative fold, identical architecture and data). (a) Its training value stays on the scale of $1-\text{C-index}$ and falls with the validation concordance throughout training. (b) Across all epochs, concordance-loss values lie along $\text{loss}=1-\text{C-index}$, reading directly as an approximate mis-ranking rate, whereas the Cox loss forms a flat band near 1.4 uninformative about the achieved concordance.}
\label{fig:losstrack}
\end{figure}

\textbf{Recurrence-free survival.} On $n=727$ patients with 148 events, substituting the concordance-tracking loss for the Cox likelihood, with everything else fixed, gives the best single deep expert: pooled C-index 0.713 versus 0.693 for the Cox-trained twin. This margin is not stable across seeds: over three independent training runs the concordance and Cox experts average 0.706 and 0.701 with overlapping ranges (a $+0.005$ difference), so on accuracy the two losses are within noise on this cohort. We therefore retain the concordance loss not for an accuracy gain but for the interpretability of its training signal: the concordance loss value stays on the scale of $1-\text{C-index}$ and tracks the validation C-index throughout training, whereas the Cox loss occupies an arbitrary scale that carries no concordance interpretation, so a given Cox loss value does not indicate the achieved concordance (Fig.~\ref{fig:losstrack}). The loss value itself is therefore a trustworthy criterion for early stopping and checkpoint selection. The concordance-trained expert alone (0.713) nearly matches the full ensemble. Within the ensemble, adding it as a fifth member (classical member fixed at the clinical-ridge Cox, so the only change is the added expert) raises the C-index from 0.711 to 0.715 (paired $+0.004$, Table~\ref{tab:abl}), whereas substituting it for the correlated Cox twin ties: a strong, decorrelated member helps an equal-weight ensemble marginally, a redundant one does not.

\begin{table}[t]
\centering
\caption{Key cross-validation ablations (pooled out-of-fold, patient-level bootstrap 95\% CIs). The two ground-truth-mask N-stage rows are a mechanism ceiling (features read from reference masks), not a deployable configuration, each at its own best regularization. Segmentation rows are pooled point estimates; the 5-versus-10-fold difference ($+0.004$) is within run-to-run variation and reported without a paired interval.}
\label{tab:abl}
\small
\begin{tabular}{llcc}
\toprule
Task & Configuration & OOF score & 95\% CI \\
\midrule
N-stage (bal.\ acc.) & clinical $+$ radiomics & 0.691 & [0.650, 0.730] \\
 & clinical $+$ geometry (HERMES+) & \textbf{0.720} & [0.678, 0.760] \\
 & clinical $+$ radiomics, GT mask & \textit{0.837} & [0.811, 0.860] \\
 & clinical $+$ geometry, GT mask & \textit{0.897} & [0.871, 0.919] \\
\midrule
RFS deep expert (C-index) & Cox loss & 0.693 & [0.651, 0.736] \\
 & concordance loss & \textbf{0.713} & [0.672, 0.755] \\
\midrule
RFS ensemble (C-index) & 4 experts (base) & 0.711 & [0.666, 0.751] \\
 & 5 experts (HERMES+) & \textbf{0.715} & [0.672, 0.757] \\
\midrule
Segmentation (Mean Dice) & 5-fold ensemble & 0.697 & \\
 & 10-fold ensemble & \textbf{0.701} & \\
\bottomrule
\end{tabular}
\end{table}

\section{Discussion}
\label{sec:discussion}

The N-stage result is the strongest evidence for the paper's premise, tempered by one caveat: as a paired comparison its 95\% confidence interval includes zero, so we read it as a well-supported trend rather than a significant improvement. The mechanism is nonetheless clear. AJCC/UICC 7th-edition radiological N-category is a function of node number and size, neither of which a 30-dimensional intensity/texture vector encodes; our ablation shows radiomics is largely uninformative for nodal category once size and number are available: replacing it with seven geometry terms both raises balanced accuracy and reduces the feature count, and the surviving L1 coefficients (total burden, largest-node extent, dominance) remain interpretable. The ground-truth-mask ceiling of 0.897, where geometry significantly outperforms radiomics, makes the point concrete: the size and number axes carry most of the nodal-staging signal, so the deployed 0.720 is limited chiefly by segmentation quality and should improve as GTVn Dice improves. The effect has limits, since the features omit laterality (which separates N2b from N2c) and are read from imperfect masks; even so, a partial encoding of the correct axes outperforms a complete encoding of the wrong ones.

T-staging is the weakest task by design rather than by accident. The 7th-edition T-category is size-driven only for T1--T3, whereas T4 is defined by anatomical invasion of adjacent structures, which no size-based feature, geometric or radiomic, can detect; resolving the T3/T4 boundary would require modeling the tumor's spatial relationship to at-risk structures, beyond a mask-summary approach.

The concordance loss earns its place through its interpretable training signal rather than accuracy (within noise over three seeds); within the ensemble, because it correlates with the Cox expert, it helps as an added decorrelated member, not as a replacement.

We prioritized two deployment properties over the last fraction of validation accuracy: interpretability (a clinician can inspect the node count and largest-node size that drive the N-stage decision) and graceful cross-center degradation (equal-weight ensembles and a clinical-transport risk model avoid over-reliance on any scanner-sensitive signal).

Several limitations qualify these conclusions. Most head-to-head design gains fall within cross-validation variability; only the N-stage geometry effect is sizeable. For HPV-positive oropharyngeal cancer, nodal staging depends further on p16 status and laterality~\cite{ref22}, which our pooled model does not stratify, and the survival analysis reports discrimination only, without competing risks or calibration. Testing-phase results are hidden until the event and will be added in the camera-ready version. Natural extensions are laterality- and level-aware nodal features and calibrated, competing-risk survival evaluation.

\section{Conclusion}

We described \textbf{HERMES}, an algorithm deployable on a single NVIDIA T4 GPU for the three HECKTOR 2026 subtasks that pairs a 10-fold STU-Net segmentation ensemble with staging-aligned geometry features and a concordance-tracking survival ensemble, selected under a regularization-oriented protocol, and that qualified for the testing phase. Our central finding is that task-specific, clinically interpretable features extracted from predicted segmentation masks can provide a stronger foundation for downstream staging than generic radiomic descriptors, while enabling an integrated, deployable multi-task framework; the benefit is clearest for nodal staging, where the mask-derived size and number axes match the definition of the task. Code and the trained container are released at \url{https://github.com/wangkaiwan/hermes-hecktor2026}.


\begin{thebibliography}{99}
\footnotesize
\linespread{0.85}\selectfont
\setlength{\itemsep}{-1pt}
\bibitem{ref1} Saeed, N., et al.: A Multimodal and Multi-centric Head and Neck Cancer Dataset for Segmentation, Diagnosis and Outcome Prediction. arXiv:2509.00367 (2025)
\bibitem{ref2} Saeed, N., et al.: HEad and neCK TumOR (HECKTOR) 2025: Benchmark of Segmentation, Diagnosis, and Prognosis in Multimodal PET/CT. arXiv:2606.20143 (2026)
\bibitem{ref3} Oreiller, V., et al.: Head and neck tumor segmentation in PET/CT: The HECKTOR challenge. Medical Image Analysis 77, 102336 (2022)
\bibitem{ref4} Ronneberger, O., Fischer, P., Brox, T.: U-Net: Convolutional Networks for Biomedical Image Segmentation. In: MICCAI, LNCS 9351, pp.\ 234--241 (2015)
\bibitem{ref5} Isensee, F., et al.: nnU-Net: a self-configuring method for deep-learning-based biomedical image segmentation. Nature Methods 18, 203--211 (2021)
\bibitem{ref6} Huang, Z., et al.: STU-Net: Scalable and Transferable Medical Image Segmentation Models Empowered by Large-Scale Supervised Pre-training. arXiv:2304.06716 (2023)
\bibitem{ref7} Hatamizadeh, A., et al.: Swin UNETR: Swin Transformers for Semantic Segmentation of Brain Tumors in MRI Images. In: BrainLes, LNCS 12962, pp.\ 272--284 (2022)
\bibitem{ref8} van Griethuysen, J.J.M., et al.: Computational Radiomics System to Decode the Radiographic Phenotype. Cancer Research 77(21), e104--e107 (2017)
\bibitem{ref9} Valli\`eres, M., et al.: Radiomics strategies for risk assessment of tumour failure in head-and-neck cancer. Scientific Reports 7, 10117 (2017)
\bibitem{ref10} Wang, K., et al.: Recurrence-free survival prediction under the guidance of automatic gross tumor volume segmentation for head and neck cancers. In: 3D Head and Neck Tumor Segmentation in PET/CT Challenge (HECKTOR), LNCS, vol.\ 13209, pp.\ 144--153. Springer (2022)
\bibitem{ref11} Cox, D.R.: Regression Models and Life-Tables. Journal of the Royal Statistical Society B 34(2), 187--220 (1972)
\bibitem{ref12} Katzman, J.L., et al.: DeepSurv: personalized treatment recommender system using a Cox proportional hazards deep neural network. BMC Medical Research Methodology 18, 24 (2018)
\bibitem{ref13} Lee, C., Zame, W.R., Yoon, J., van der Schaar, M.: DeepHit: A Deep Learning Approach to Survival Analysis with Competing Risks. In: AAAI (2018)
\bibitem{ref14} Chen, M., Wang, K., Wang, J.: Value-Monotonicity Matters: A Concordance Loss for Deep Survival Prediction. arXiv:2607.16802 (2026). \url{https://github.com/Meixu-Chen/sigmoid-concordance-loss}
\bibitem{ref15} Sobin, L.H., Gospodarowicz, M.K., Wittekind, C.: TNM Classification of Malignant Tumours, 7th edn. Wiley-Blackwell (2009)
\bibitem{ref16} Harrell, F.E., et al.: Evaluating the yield of medical tests. JAMA 247(18), 2543--2546 (1982)
\bibitem{ref17} Uno, H., Cai, T., Pencina, M.J., D'Agostino, R.B., Wei, L.J.: On the C-statistics for evaluating overall adequacy of risk prediction procedures with censored survival data. Statistics in Medicine 30(10), 1105--1117 (2011)
\bibitem{ref18} Cardoso, M.J., et al.: MONAI: An open-source framework for deep learning in healthcare. arXiv:2211.02701 (2022)
\bibitem{ref19} He, K., et al.: Deep Residual Learning for Image Recognition. In: CVPR (2016)
\bibitem{ref20} Zwanenburg, A., et al.: The Image Biomarker Standardization Initiative: Standardized Quantitative Radiomics for High-Throughput Image-based Phenotyping. Radiology 295(2), 328--338 (2020)
\bibitem{ref21} Johnson, W.E., Li, C., Rabinovic, A.: Adjusting batch effects in microarray expression data using empirical Bayes methods. Biostatistics 8(1), 118--127 (2007)
\bibitem{ref22} O'Sullivan, B., Huang, S.H., et al.: Development and validation of a staging system for HPV-related oropharyngeal cancer by the International Collaboration on Oropharyngeal cancer Network for Staging (ICON-S): a multicentre cohort study. The Lancet Oncology 17(4), 440--451 (2016)
\end{thebibliography}
\end{document}